\documentclass[journal]{IEEEtran}
\usepackage{amssymb}
\usepackage[subfigure]{graphfig}
\usepackage{amsmath}
\usepackage{amsfonts}
\usepackage{graphicx}
\usepackage{epsfig}
\usepackage{cases}
\usepackage{colortbl}
\usepackage{array,color}
\usepackage{algorithm}
\usepackage{algorithmic}
\usepackage{multirow}
\usepackage{color}

\usepackage{pifont}
\usepackage{multirow}

\usepackage{array}
\newcolumntype{I}{!{\vrule width 3pt}}
\newlength\savedwidth

\newlength\savewidth
\newcommand\shline{\noalign{\global\savewidth\arrayrulewidth
                            \global\arrayrulewidth 1.0pt}%
                   \hline
                   \noalign{\global\arrayrulewidth\savewidth}}

\ifCLASSINFOpdf
\else
\fi
\begin{document}
%
\title{Improving Neural Radiance Fields with Depth-aware Optimization for Novel View Synthesis}
%
%
%

\author{Shu~Chen,
       Junyao~Li,
        Yang~Zhang,
        and~Beiji~Zou
\thanks{S.~Chen, J.~Li and Y.~Zhang are with the School of Computer Science.School of Cyberspace Security, Xiangtan University, Xiangtan, China and Key Laboratory of Intelligent Computing \& Information Processing, Ministry of Education, Xiangtan, China (e-mail: csu\_cs@163.com; 351083542@qq.com; 460086012@qq.com.}
\thanks{B. Zou is with School of Computer Science and engineering, Central South University, Changsha, China (e-mail: bjzou@csu.edu.cn).}}
\maketitle

\begin{abstract}
With dense inputs, Neural Radiance Fields (NeRF) is able to render photo-realistic novel views under static conditions. Although the synthesis quality is excellent, existing NeRF-based methods fail to obtain moderate three-dimensional (3D) structures. The novel view synthesis quality drops dramatically given sparse input due to the implicitly reconstructed inaccurate 3D-scene structure. We propose SfMNeRF, a method to better synthesize novel views as well as reconstruct the 3D-scene geometry. SfMNeRF leverages the knowledge from the self-supervised depth estimation methods to constrain the 3D-scene geometry during view synthesis training. Specifically, SfMNeRF employs the epipolar, photometric consistency, depth smoothness, and position-of-matches constraints to explicitly reconstruct the 3D-scene structure. Through these explicit constraints and the implicit constraint from NeRF, our method improves the view synthesis as well as the 3D-scene geometry performance of NeRF at the same time. In addition, SfMNeRF synthesizes novel sub-pixels in which the ground truth is obtained by image interpolation. This strategy enables SfMNeRF to include more samples to improve generalization performance. Experiments on three public datasets demonstrate that SfMNeRF surpasses state-of-the-art approaches given the spare inputs. Code is available at {\color{blue}https://github.com/XTU-PR-LAB/SfMNeRF.}
\end{abstract}


%
\IEEEpeerreviewmaketitle

\section{Introduction}
%
%
%
%
\IEEEPARstart{N}{eural} Radiance Fields \cite{Mildenhall2020} have shown very impressive results for novel-view synthesis. NeRF employ a continuous five-dimensional (5D) function to implicitly encode the three-dimensional structure and appearance of a specific scene which is represented by a training multi-layer perceptron (MLP), and the novel view of the scene is synthesized by volumetric rendering. NeRF explicitly constrain the synthesized pixels to close the ground truth in which the employed volumetric rendering enables NeRF to implicitly constrain the 3D-scene geometry.  However, the emitted colors and volume densities are entangled in the NeRF so that the NeRF require dense input views to eliminate the geometric ambiguity. When the input views are sparse, NeRF are prone to finding a degenerate solution to the image reconstruction objective \cite{ZhangK2020}. On the other hand, NeRF fail to synthesize good novel views in the scenes with large texture-less regions such as couches and floors because it is unlikely to acquire sufficient correspondences cross view in NeRF.

The poor performance of NeRF in the sparse inputs or the scenes containing many texture-less regions comes from the implicitly estimated inaccurate 3D shape. To overcome this problem, some works leverage depth priors to improve the novel-view synthesis performance of neural radiance fields. These depth data are commonly estimated by running a structure-from-motion (SfM) approach \cite{Ye2024, Wang2012} from the input images. DS-NeRF \cite{DengK2022} adopts the sparse 3D points from a SfM as the supervision in the NeRF optimization. However, the depth priors from SfM are commonly sparse and noisy. To obtain dense depth priors, NerfingMVS \cite{WeiY2021} employs a monocular depth network with the sparse depth from SfM reconstruction as the supervision to obtain the depth priors, and then using the adapted depth priors to guide volume sampling in the optimization of NeRF. Roessle et al. \cite{Roessle2022} propose a similar approach, they adopt depth completion to convert these sparse points into dense depth maps, which are used to guide NeRF optimization. However, this kind of solution still treats depth estimation and view synthesis as two separate processes that cannot benefit from each other.

Inspired by the self-supervised depth estimation approaches \cite{ZhouT2017, ChenS2022}, we integrate the novel-view synthesis and depth prediction into a single end-to-end procedure so they can benefit from each other. The employed explicit depth estimation can compensate for the weakly implicit 3D-geometry constraint in NeRF. With the improvement of depth estimation, the quality of novel-view synthesis is enhanced accordingly. Specifically, we employ the epipolar geometry to eliminate the pixels' depth uncertainty by constraining the corresponding points in another image of one point in one image must to be on a ray called the epipolar line. Furthermore, we leverage a patch photometric consistent loss across multiple views in which the image warping is implemented to ensure the identical region in different views has the same appearance. To further reduce the 3D-shape ambiguity, the surface smooth constraint in scenes and the position-of-matches constraint that the identical feature in different views has the same world coordinates are implemented. At the same time, we implement sub-pixel rendering in which the 2D image coordinates of pixels to be synthesized are not integer vectors but float vectors that are sampled based on the coordinates of image pixels with (0, 1) offset. The colors of the sampled sub-pixels are obtained by bilinear interpolation on the images. This new sampling strategy enables SfMNeRF to include more samples to improve generalization performance.

To sum up, our main contributions include:

1) By employing the cross-view consistent patch-based photometric loss, SfMNeRF explicitly constrains the 3D-scene geometry that reduces the geometric ambiguity to some extent in NeRF.

2) The employed epipolar, smooth and position-of-matches constraints enable SfMNeRF to be aware of the scene's structure, so the quality of novel-view synthesis improves.

3) The implemented sub-pixel rendering in SfMNeRF improves the generalization ability of the NeRF.

\section{Related Work}
We briefly review depth-aware NeRF and self-supervised depth prediction in this section.
\subsection{Depth-aware NeRF}
NeRF have achieved the best results for novel-view synthesis under static conditions. NeRF implicitly represent a scene as a continuous five-dimensional (5D) function by training a MLP and then use volume rendering to synthesize a novel view. Many NeRF's variants have been proposed, such as fast training and inference \cite{Muller2022, YuA2022}, modelling non-rigid scenes \cite{Pumarola2021}, scalable \cite{Tancik2022}, unbounded scenes \cite{ZhangK2020-2, Barron2022}, editable \cite{YangB2022, LiuS2021}, handling reflections \cite{Verbin2022} and generalization \cite{YuA2021, XuQ2022}. Although NeRF can achieve better performance for novel-view synthesis under rich texture and dense image inputs, they synthesize unplausible results due to the inherited geometric ambiguity when the scene is observed by sparse. Some works enforce regularization to improve the novel-view synthesis performance when the input is sparse, for example, by  penalizing a semantic consistency loss \cite{JainA2021} or introducing extra unobserved viewpoints \cite{Niemeyer2022}. However, existing approaches either regularize the depth with estimated sparse 3D point clouds from a SfM, or heavily rely on a extra multi-view dataset that might hard to collect or not be available. In contrast, our approach introduces more samples by sub-pixel rendering to enhance the generalization ability of the NeRF. Depth priors are also employed as the supervision in the NeRF optimization to guarantee a unique estimated 3D-scene geometry to handle the artifacts in sparse input scenarios. Here, we review the depth-aware Neural Radiance Fields in both scene-level and cross-scenes-level.

Depth-aware NeRF are expected to learn an accurate 3D shape to improve the novel-view synthesis performance \cite{DengK2022, WeiY2021, Roessle2022, WangQ2021}. DS-NeRF \cite{DengK2022} employs the sparse depth information from COLMAP \cite{Schonberger2016} as the supervision to optimize the depth value of the pixel rendered. NerfingMVS \cite{WeiY2021} proposes a similar approach, but the depth priors are obtained by finetuning a monocular depth neural network on its sparse SfM reconstruction from the target scene. Dense Depth Priors \cite{Roessle2022} uses a depth completion network to convert the sparse depth data of each view obtained from the SfM into dense depth maps individually, which are used to guide NeRF optimization and supervise the depth of the pixel rendered. However, Dense Depth Priors process each view individually that it is not view-consistent. The depth priors in these approaches are commonly pre-obtained by an independent approach and the view synthesis and depth estimation are separated so that they cannot benefit from each other. In comparison, our approach estimates the 3D-structure of the scene and synthesizes the novel view at the same time.

The per-scene NeRF optimize the representation of each scene individually, so it is time consuming and lack of generalization. To resolve these shortcomings, prior works also present generalizable radiance-field-based methods \cite{YuA2021, WangQ2021, Trevithick2021}. pixelNeRF \cite{YuA2021} introduces a fully convolutional architecture to extract the feature maps of input images which are conditioned on the NeRF to learn a scene prior by training across multiple scenes. However, the features in pixelNeRF are aggregated from a multi-view 2D image. Point-NeRF \cite{XuQ2022} leverages deep multi-view stereo (MVS) techniques to reconstruct point clouds which are represented as anchors for feature extraction.

\subsection{Self-supervised Depth Prediction}
Self-supervised depth prediction approaches aim to predict depth directly from monocular images, by enforcing a photometric loss on corresponding stereo images \cite{Godard2017} or on temporally adjacent frames \cite{ZhouT2017}. This kind of approach typically synthesizes a new view by image warping to serve as the supervisory signal. To eliminate the adverse effects of dynamic objects or occlusions, a mask explained for the motion is introduced to ignore certain regions that do not satisfy the static scene assumption. Zhou et al. \cite{ZhouT2017} trained two networks, one for mask and another for depth estimation, to reconstruct the depth of scene from monocular images. In practice, the predicted mask is inaccurate that introduces more errors during training. Instead of being learned from a network, Godard et al. \cite{Godard2017} introduce an auto-mask to eliminate training pixels that violate camera motion assumptions. Other methods alleviate this problem by either modeling the motion of individual objects \cite{Casser2019, YinZ2018}, estimating dense 3D translation field \cite{LiH2020}, or filtering by instance segmentation \cite{Lee2020}. Garg et al. \cite{Garg2016} introduce photometric consistency constraint to recover depth from stereo pairs, which is further improved by left-right consistency constraints \cite{Godard2017}. Furthermore, some works \cite{LiR2018, ZhanH2018} use both temporal and spatial photometric warp errors to train the model. Inspired by self-supervised learning methods, in this work, we employed the patch-based multi-view consistent photometric constraint to guarantee an accurate 3D-scene's shape.

\section{Preliminary}
A neural radiance field is represented by a continuous 5D function \emph{f} that maps a 3D coordinate ${\bf{x}} = \left( {x,y,z} \right)$ and viewing direction $\left( {\theta ,\varphi } \right)$ to a volume density $\sigma $ and an emitted color ${\bf{c}}$. The continuous function is implicitly parameterized by a multi-layer perceptron, and the weights of the multi-layer perceptron are optimized to synthesize the input images of a specific scene.
\begin{equation}\label{eq1}
{f_w }:\left( {\beta \left( {\bf{x}} \right),\beta \left( {\bf{d}} \right)} \right) \to \left( {{\bf{c}},\sigma } \right),
\end{equation}
where $w$ denotes the network weights, and $\beta$ represents a predefined positional encoding applied to ${\bf{x}}$ and ${\bf{d}}$.

Given \emph{m} training images and the corresponding camera poses, a photometric loss is leveraged to optimize the NeRF as
\begin{equation}\label{eq2}
L = \frac{1}{m}\sum\limits_{i = 1}^m {\left\| {{I_i} - {{\hat I}_i}} \right\|_2^2} ,
\end{equation}
where ${I_i}$ and ${\hat I_i}$ are the ground-truth color of image \emph{i} and the corresponding synthesized image by volume rendering, respectively.

For each pixel of ${\hat I_i}$, casting a ray ${\bf{r}}\left( t \right) = {\bf{o}} + t{\bf{d}},{\rm{ }}{\bf{o}} \in {\Re ^3},{\rm{ }}{\bf{d}} \in {S^2},{\rm{ }}t \in \left[ {{t_n},{t_f}} \right]$ from the camera center ${\bf{o}}$ through the pixel along direction ${\bf{d}}$, and its color ${{\bf{\hat c}}_\theta }$ is computed using alpha compositing:
\begin{equation}\label{eq3}
{{\bf{\hat c}}_\theta }\left( {\bf{r}} \right) = \int_{{t_n}}^{{t_f}} {T\left( t \right){\sigma _\theta }\left( {{\bf{r}}\left( t \right)} \right)} {{\bf{c}}_\theta }\left( {{\bf{r}}\left( t \right),{\bf{d}}} \right)dt,
\end{equation}
where $T\left( t \right) = \exp \left( { - \int_{{t_n}}^t {{\sigma _\theta }\left( {{\bf{r}}\left( s \right)} \right)ds} } \right),$ and ${\sigma _\theta }\left(  \cdot  \right)$ and ${{\bf{c}}_\theta }\left( { \cdot , \cdot } \right)$ indicate the volume density and color prediction of the radiance field, respectively.

\section{SfMNeRF}
\subsection{Overview}
Figs. 1 and 2 show the pipeline of our SfMNeRF framework for view synthesis. Firstly, a set of two images was selected from the dataset. In the set, one image is assumed to be the reference image, and another image selected is this that has an overlapping region with the reference image. Secondly, we extracted scale-invariant feature transform (SIFT) features in each image and obtained the matched SIFT correspondences between them. Finally, the matched SIFT correspondences were input into the neural radiance field to obtain their 3D coordinates. We used the positions of matched features loss to guarantee an accurate geometric shape based on the fact that the identical features in different views have the same world coordinates. Furthermore, we employed the epipolar constraint to enforce the correspondence of one point in reference image must lie in the epipolar line in another image by optimizing the minimum of the difference between the 3D coordinates of the point and the epipolr points. In addition, we leveraged the patch-based multi-view consistent photometric constraint to further constrain the estimated 3D-scene structure. We randomly selected a patch from the reference image and added random offset between (0, 1) to the image coordinates of each pixel in the patch to obtain sub-pixels; then the coordinates of sub-pixels $(x,y)$ and the direction ${\bf{d}}$ were input into the neural radiance field which output the color and depth of each sub-pixel. The ground-truth of each sub-pixel was obtained by a bilinear interpolator from the reference image. The pixels in the patch of the reference image were warped into the other view to synthesize a new patch, and a photometric reconstruction loss between the patch in the reference image and the new synthesized patch was implemented to constrain the estimated depth of each sub-pixel from the neural radiance field. In addition, the depth smooth loss was employed to further constrain the estimated 3D-scene geometry. To prevent getting stuck in local minima, we implement SfMNeRF in a pyramidal strategy.

\begin{figure*}
  \centering
  \includegraphics[width = 6.5in]{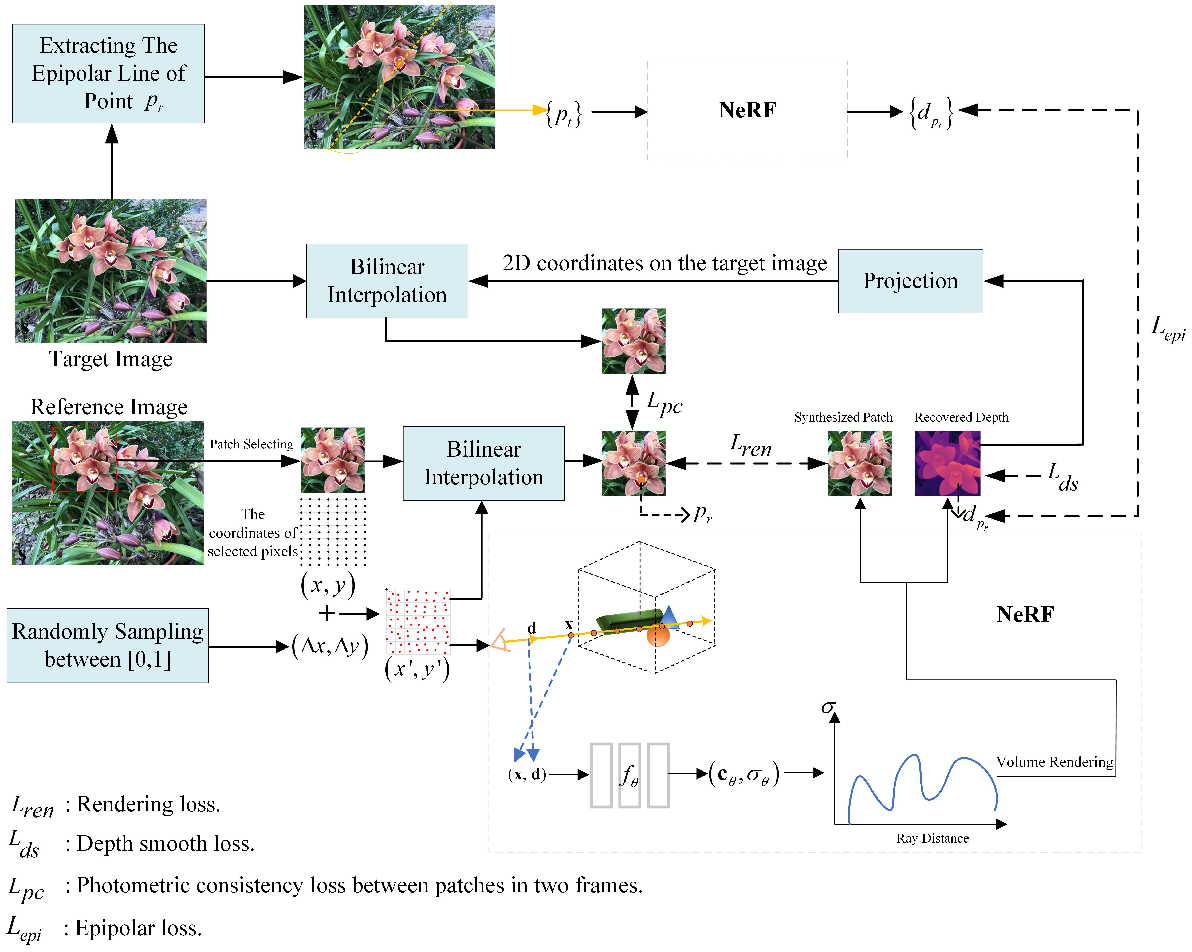}\\
  \caption{Overview of SfMNeRF. This picture depicts how the photometric consistency loss and epipolar loss are implemented.
}\label{f1}
\end{figure*}

\begin{figure*}
  \centering
  \includegraphics[width = 6.0in]{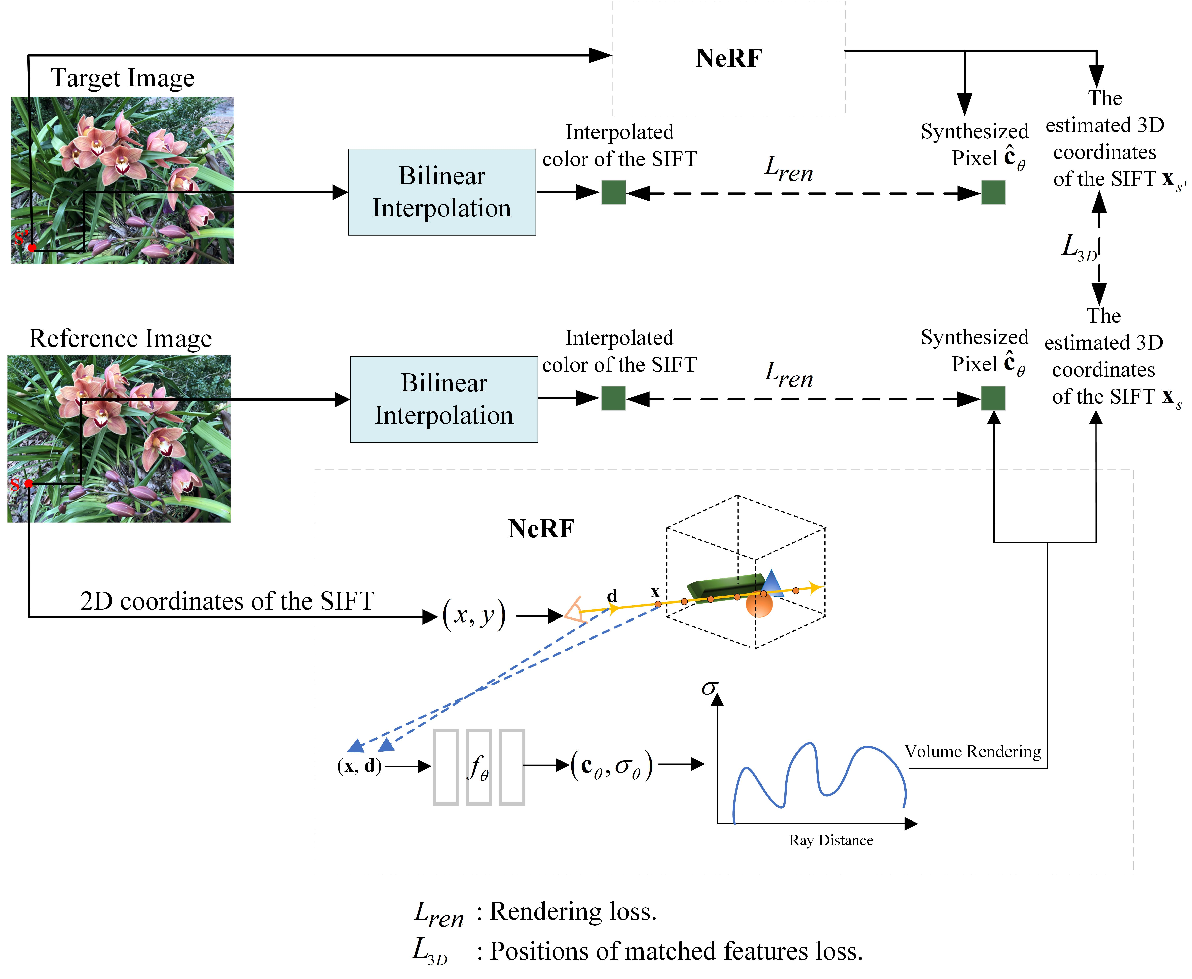}\\
  \caption{Overview of SfMNeRF. This picture depicts how the positions of matched features constraint is implemented. $S$  and $S'$ are the matched features, respectively.
}\label{f1}
\end{figure*}

\begin{figure}[!t]
  \centering
  \includegraphics[width = 3.8in]{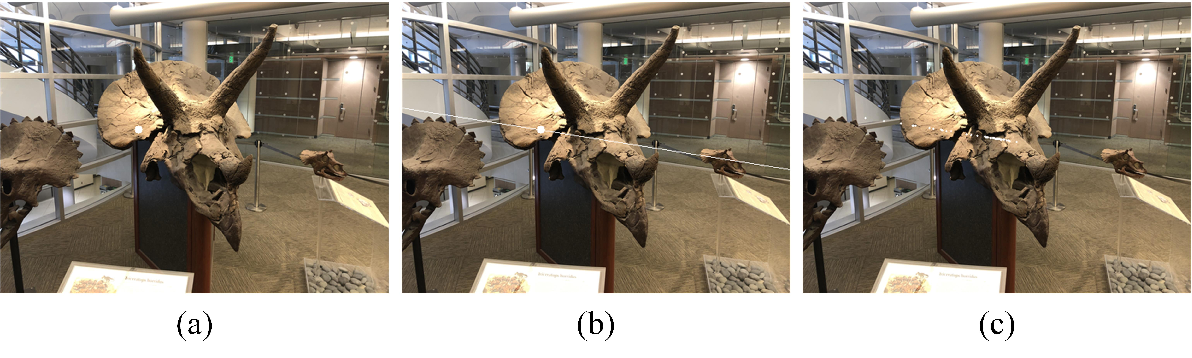}\\
  \caption{Implausible epipolar points elimination. (a) The reference image with a white point. (b) The epipolar line is depicted by the white line in another image and the white point is the corresponding point. (c) The obtained epipolar points after filtering which represented by the white points in the image.
}\label{f3}
\end{figure}
\begin{figure*}[!t]
  \centering
  \includegraphics[width = 6.5in]{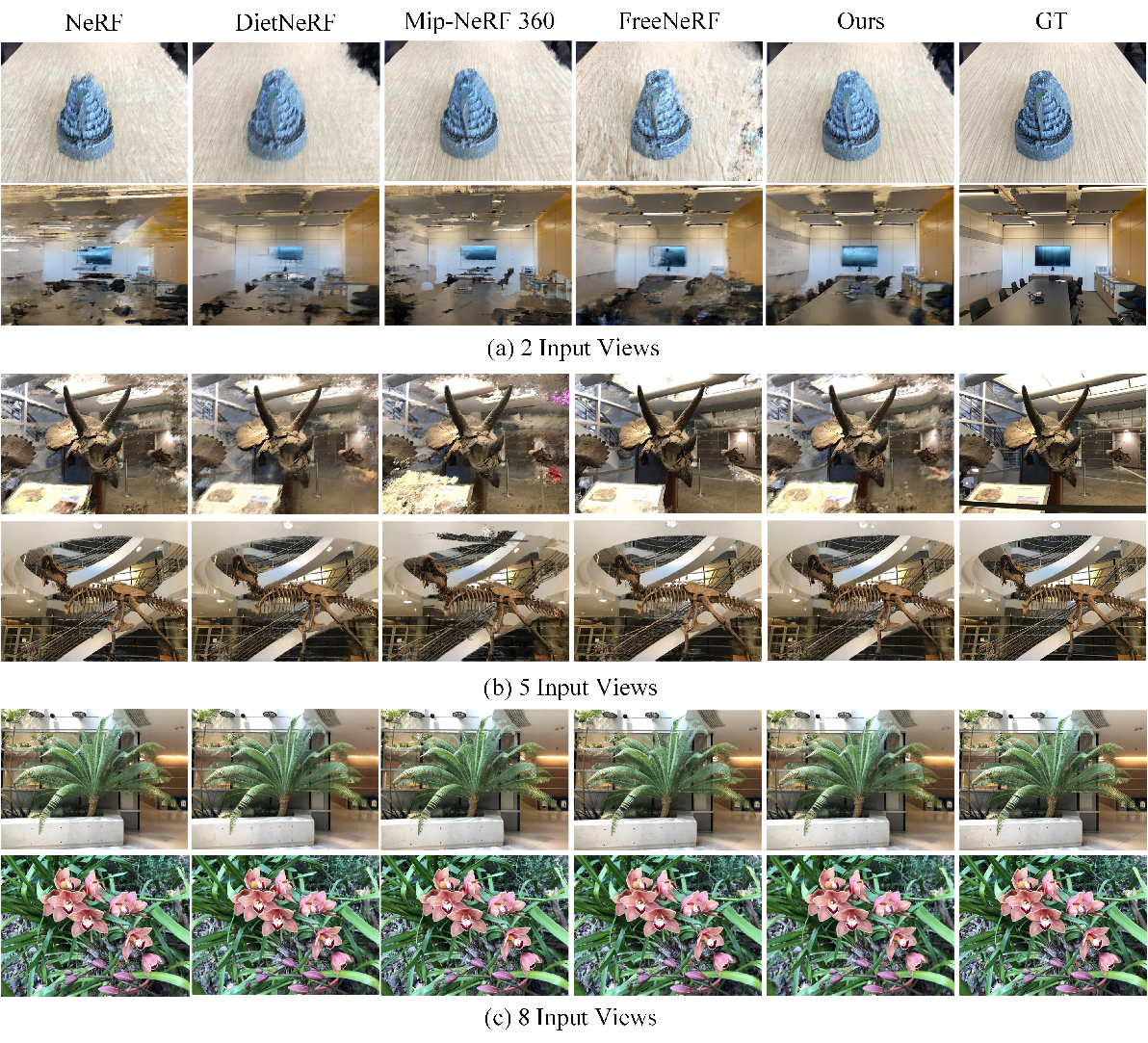}\\
  \caption{Qualitative comparison between our SfMNeRF and other approaches on the LLFF-NeRF dataset.
}\label{f4}
\end{figure*}
\begin{figure*}[!t]
  \centering
  \includegraphics[width = 6.5in]{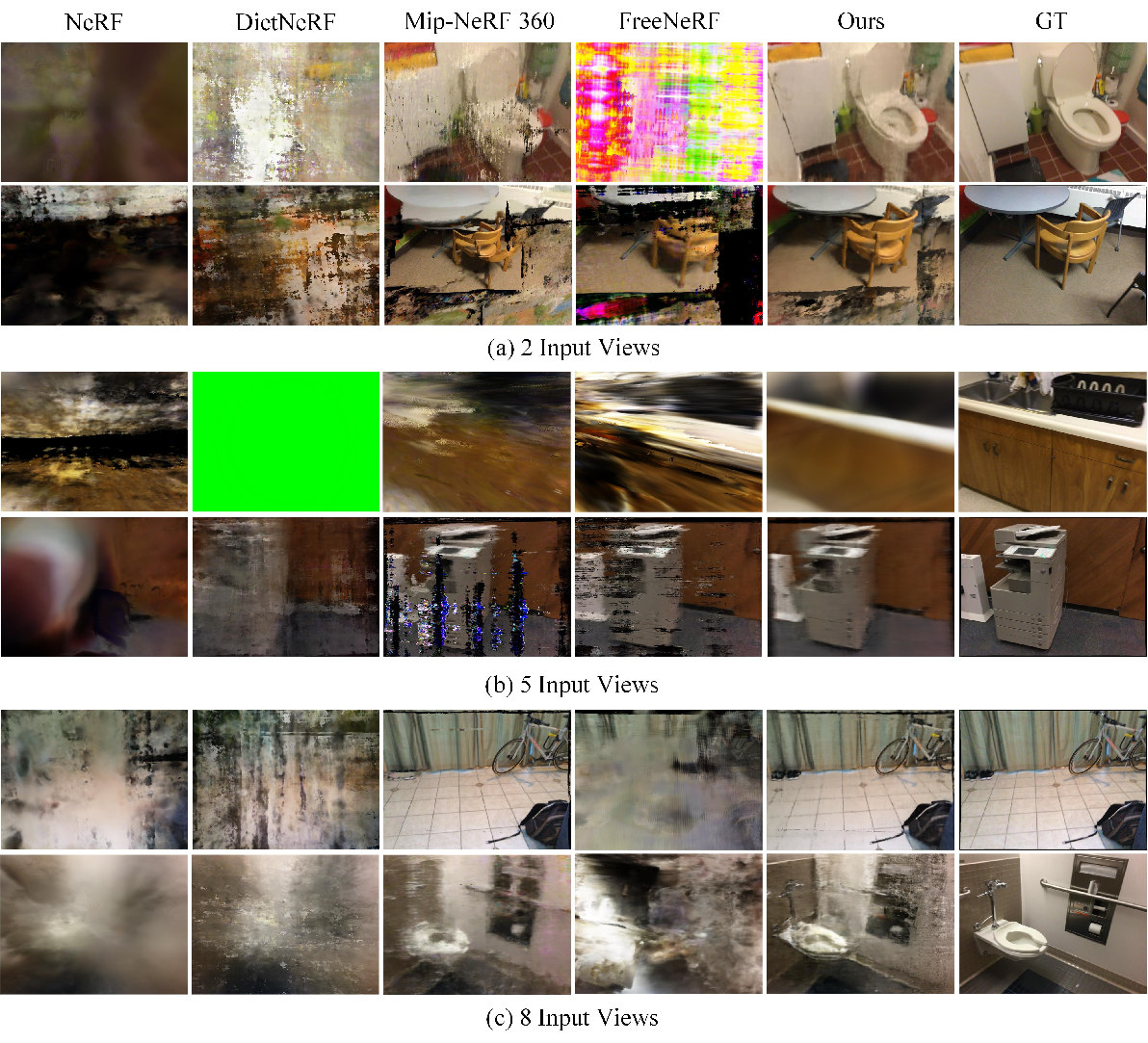}\\
  \caption{Qualitative comparison between our SfMNeRF and other approaches on the ScanNet dataset.
}\label{f5}
\end{figure*}
\begin{figure*}[!t]
  \centering
  \includegraphics[width = 6.5in]{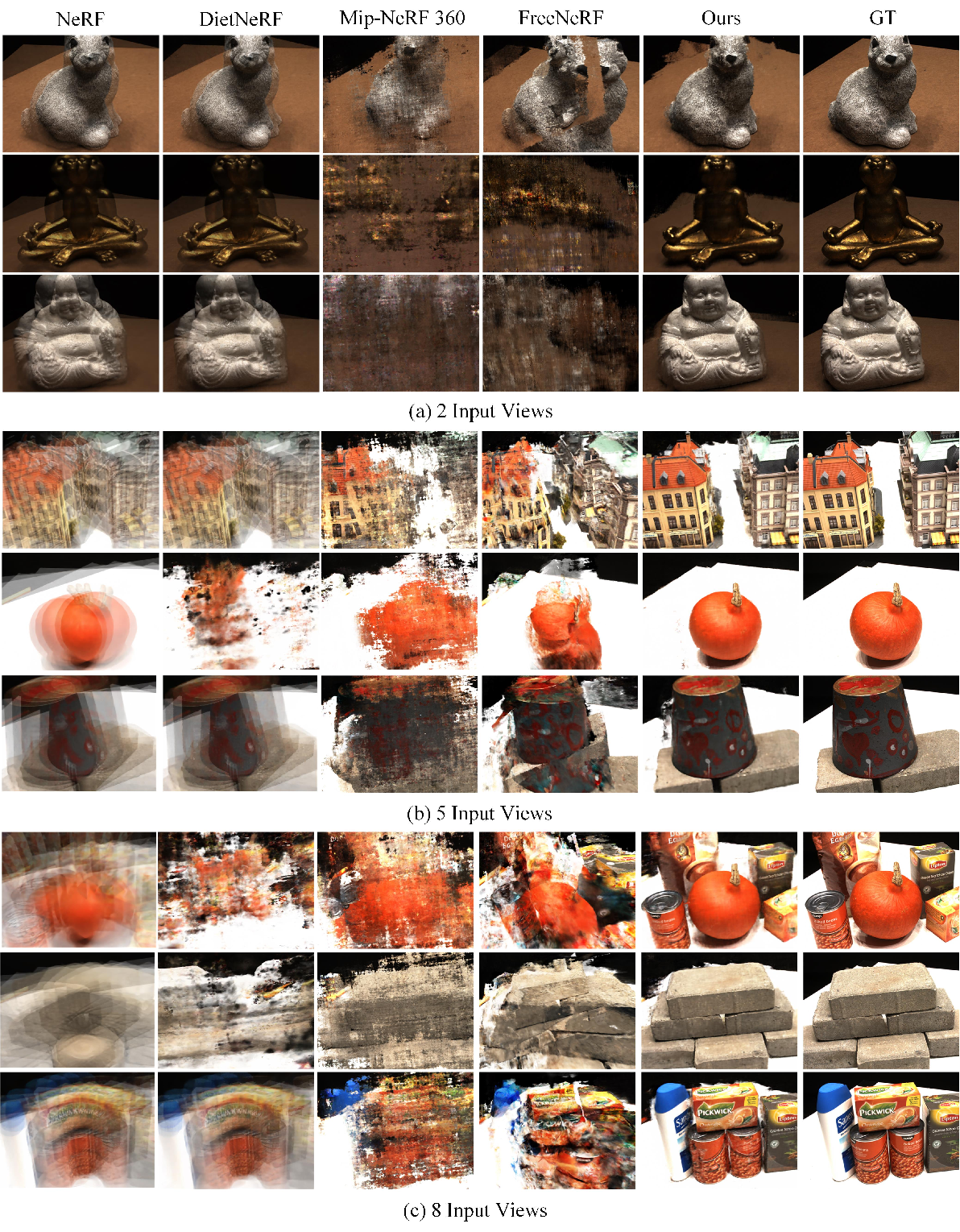}\\
  \caption{Qualitative comparison between our SfMNeRF and other approaches on the DTU dataset.
}\label{f6}
\end{figure*}

\begin{figure*}[!t]
  \centering
  \includegraphics[width = 6.5in]{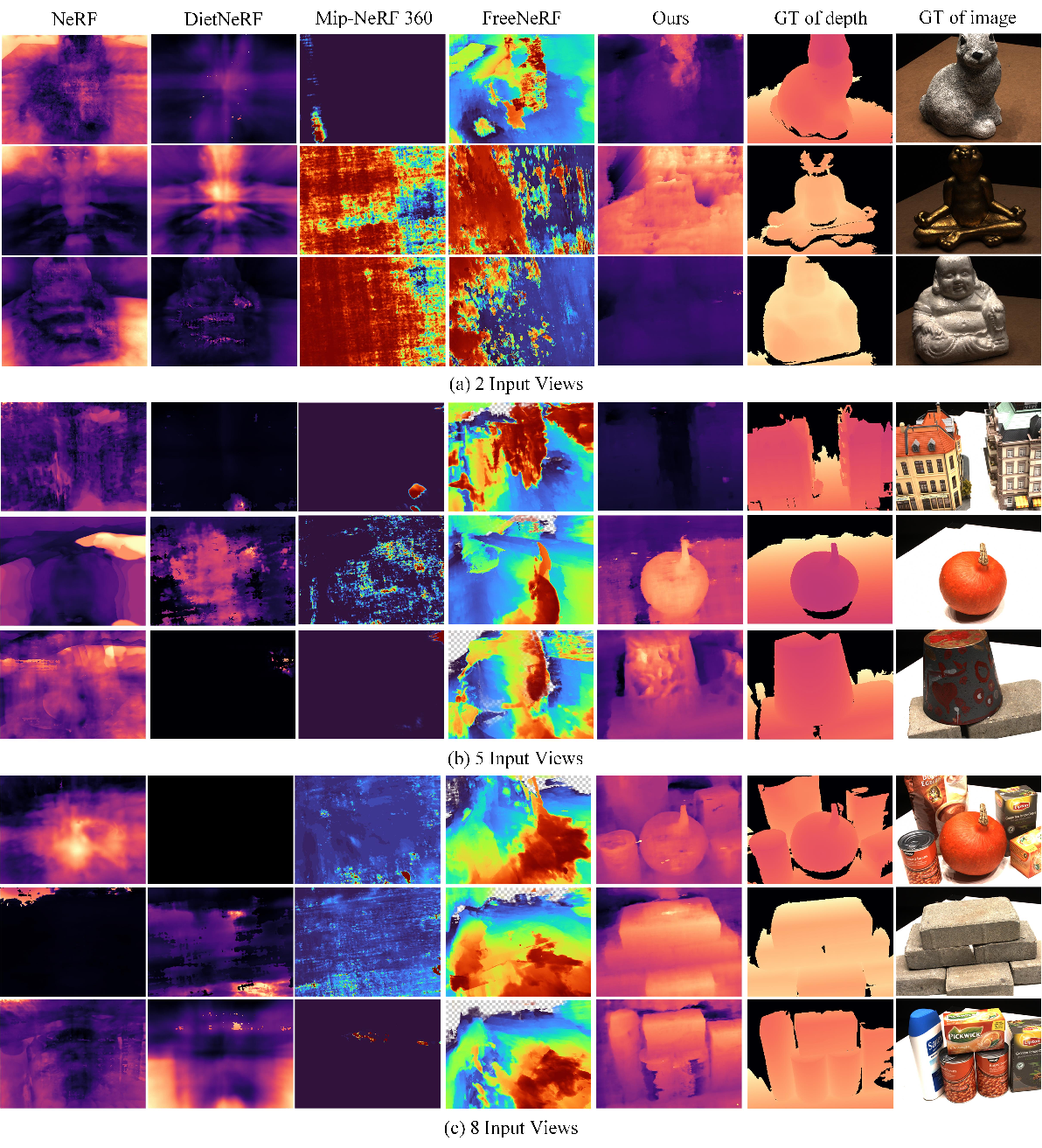}\\
  \caption{Estimated depth maps for test views in Fig. 6.
}\label{f7}
\end{figure*}

\subsection{Sub-pixel Rendering}
To achieve sub-pixel rendering, we first sample a patch from the reference image at random; the sampling region is denoted as ${\Omega _p}$. For any point $p = \left( {x,y} \right) \in {\Omega _p}$ in the patch, we add an offset to it as
\begin{equation}\label{eq4}
p = \left( {x + {x_{off}},y + {y_{off}}} \right)\left| {{x_{off}} \in \left( {0,1} \right),{y_{off}} \in \left( {0,1} \right)} \right..
\end{equation}

The ground-truth color of the sub-pixel is obtained by linearly interpolating the color values of the four-pixel neighbors (top-left, top-right, bottom-left, and bottom-right) as
\begin{equation}\label{eq5}
\hat I(p) = \sum\nolimits_{i \in \left\{ {top,bottom} \right\},j \in \left\{ {left,right} \right\}} {{w^{ij}}} I\left( {{p^{ij}}} \right),
\end{equation}
where ${w^{ij}}$ is determined according to linearly proportional to the Euclidean distance between $p$ and ${p^{ij}}$, and $\sum\nolimits_{i,j} {{w^{ij}}}  = 1$.

According to NeRF [1], the rendering loss is formulated as
\begin{equation}\label{eq6}
{L_{ren}} = \frac{1}{m}\sum\limits_{{p_i} \in {\Omega _p}} {\left\| {\hat I\left( {{p_i}} \right) - \tilde I\left( {{p_i}} \right)} \right\|_2^2} ,
\end{equation}
where $\hat I\left( {{p_i}} \right)$ is the ground-truth color of point ${p_i}$ and $\tilde I\left( {{p_i}} \right)$ is the corresponding synthesized pixel by volume rendering.

\subsection{Positions of Matched Features Constraint}
As introduced in SaNeRF \cite{ChenS2022-2}, we also leverage the 3D coordinates of sparse keypoints to guide the geometry optimization in the NeRF, based on the fact that the identical keypoints in different views have the same world coordinates.

The 3D coordinates of the SIFT features are formulated as a weighted sum of all the samples volume densities ${\sigma _j}$ output from the NeRF along the ray, defined as
\begin{equation}\label{eq7}
{{\bf{x}}_s} = \sum\limits_{j = 1}^{{N_c}} {{w_j}\left( {{\bf{o}} + {t_j}{\bf{d}}} \right)} ,{\rm{ }}{w_j} = {T_j}\left( {1 - \exp \left( { - {\sigma _j}{\delta _j}} \right)} \right),
\end{equation}
where ${\delta _j} = {t_{j - 1}} - {t_j}$ represents the distance between two consecutive samples and $N_c$ denotes the number of samples.

We denote the 3D coordinates of matched correspondences as $\left\{ {\left[ {{\bf{x}}_s^k,{\bf{x}}_{s'}^k} \right]\left| {k = 1, \cdots ,m} \right.} \right\}$, and the positions-of-matched features loss is formulated as

\begin{equation}\label{eq8}
{L_{3D}} = \frac{1}{n}\sum\limits_{k = 1}^n {\left\| {{\bf{x}}_s^k - {\bf{x}}_{s'}^k} \right\|_2^2} ,
\end{equation}
where ${\bf{x}}_s^k$ and ${\bf{x}}_{s'}^k$ are the estimated 3D coordinates of the matched SIFT features \emph{k} in the reference image and the target image according to (7), respectively.

\subsection{Epipolar Constraint}
Given the pose between two images, we can calculate the fundamental matrix. According to epipolar geometry, given one point in the reference image, we can derive a ray called epipolar line from the estimated fundamental matrix, the corresponding points in the target image must to be on the epipolar line (depicted as the yellow dot line) as shown in the Figure 1. In this work, we use the epipolar constraint to regularize the predicted depth of points from NeRF.

Given a point in the reference image, ${p_r}$, the 3D coordinates of this point is denoted as ${d_{p_r}}$, and the points in the epipolar line are denoted as $\left\{ {{p_t}} \right\}$, we define the epipolar constraint as
\begin{equation}\label{eq9}
{L_{epi}} = \mathop {\min }\limits_t \left\{ {\left\| {{d_{p_r}} - {d_{p_t}}} \right\|_2^2} \right\},
\end{equation}
where ${d_{{p_r}}}$ and ${d_{{p_t}}}$ are estimated according to (7).

Actually, the number of epipolar points is very large, it is necessary to eliminate the implausible points to improve the computational efficiency. On the other hand, some points may occluded that it is not possible to find the corresponding points in the epipolar line. We use the color similarity to eliminate the implausible points, only the points in the epipolar line in which the color difference between them and the point in the reference image below a threshold are filtered as the epipolar points. We show the filtering visual results in the Fig. 3.

\subsection{Photometric Consistency Constraint between Multiple Views}
To eliminate the shape-radiance ambiguity in the NeRF \cite{Mildenhall2020}, for each view, we explicitly leverage the patch-based photometric consistency constraint between multiple views to constrain the depth value of every pixel in the patch.

Firstly, the 3D coordinates of each pixel \emph{p} in the patch are estimated by (7) and denoted as ${{\bf{x}}_p}$. Secondly, the pixel in the reference image is warped by a back-project process onto the target image (denoted as ${I_t}$) as the following equation.
\begin{equation}\label{eq10}
{\hat p_t} = K{T_{r -  > t}}{{\bf{x}}_p},
\end{equation}
where $K$ is camera intrinsic matrix, and ${T_{r -  > t}} = \left[ {{\bf{r}},{\bf{t}}} \right]$ represents the relative pose between the reference image and the target image and defined as
\begin{equation}\label{eq11}
{\bf{r}} = {{\bf{r}}_t}^{ - 1},{\bf{t}} = - {{\bf{r}}_t}^{ - 1}{{{\bf{t}}_t}},
\end{equation}
where $\left[ {{{\bf{r}}_t},{{\bf{t}}_t}} \right]$ is the global pose of the target image.

Similar to (5), the color of the projected point ${\hat p_t}$ is obtained by a linear interpolator and denoted as $\hat I_t({\hat p_t})$.

Finally, the patch-based photometric consistency constraint is formulated as the combination of a photometric reconstruction loss ${L_{pr}}$ and a structured similarity (SSIM) loss ${L_{SSIM}}$, and defined as ${L_{pc}} = {L_{pr}} + {L_{SSIM}}$.

The photometric reconstruction loss is defined as

\begin{equation}\label{eq12}
\begin{array}{l}
{L_{pr}} = \sum\limits_{p\left( k \right) \in {\Omega _p}} {M\left( {p\left( k \right)} \right)} \left| {{{\hat I}_r}\left( {p\left( k \right)} \right) - {{\hat I}_t}\left( {{{\hat p}_t}\left( k \right)} \right)} \right|,
\end{array}
\end{equation}
where ${\hat I_r}\left( {p\left( k \right)} \right)$ is the pixel value of point \emph{k} in the patch in the reference image, and ${\hat I_t}\left( {{{\hat p}_t}\left( k \right)} \right)$ is the synthesized pixel value of projected corresponding one in the target image. $M\left( . \right)$ represents  the mask map for the patch to filter the pixels which do not have the correspondences in the target image. We pre-established a minimum rectangle to include all matched SIFT features in the reference image, if any point $p\left( k \right)$ in the patch is in the minimum rectangle where its mask $M\left( {p\left( k \right)} \right)$ is set as one, otherwise set as zero.

The SSIM loss is defined as
\begin{equation}\label{eq13}
{L_{SSIM}} = M\left( {\frac{{1 - SSI{M_{r -  > t}}}}{2}} \right),
\end{equation}
where $SSI{M_{r -  > t}}$ is the structured similarity between the patch in the reference image and the target image. $M$ is a binary mask that is set as one if the patch is in the minimum rectangle, otherwise it is set as zero.

\subsection{Depth Smooth loss}
Similar to \cite{ChenS2022}, we enforce the depth smooth prior to the estimated depth maps in the NeRF [1]. Since large image gradients commonly mean depth discontinuities, we consider the edge constraint in our approach by enforcing the L1 norm of the depth gradients in the total loss, which is formulated as the weighted image gradients across adjacent pixels.
\begin{equation}\label{eq14}
{L_{ds}} = \sum\limits_{p\left( k \right) \in {\Omega _p}} {\left| {\nabla D\left( {p\left( k \right)} \right) \cdot \left( {{e^{ - \left| {\hat I\left( {p\left( k \right)} \right)} \right|}}} \right)} \right|} ,
\end{equation}
where $\nabla$ represents the 2D differential operator and $D\left( {p\left( k \right)} \right)$ is the estimated depth of pixel $p\left( k \right)$.

\subsection{Training loss}
The total loss is formulated as a combination of the aforementioned losses; each loss is controlled by a factor.
\begin{equation}\label{eq15}
\begin{split}
{L_{total}} = {\lambda _{ren}}{L_{ren}} + {\lambda _{3D}}{L_{3D}} + {\lambda _{pr}}{L_{pr}} \\
+ {\lambda _{epi}}{L_{epi}} + {\lambda _{SSIM}}{L_{SSIM}} + {\lambda _{ds}}{L_{ds}}.
\end{split}
\end{equation}

\subsection{Pyramidal implementation of SfMNeRF}
In SfMNeRF, the photometric consistency loss is estimated based on the gradient of pixel color that constrains the patch in the target image can not be far from the patch in the reference image, otherwise, it will stuck in the local minimum. To prevent getting stuck in local minima, we implement SfMNeRF in a pyramidal strategy to handle large pixel motions. First, we construct the \emph{L}-levels pyramid representations of the reference image and the target image as $\left\{ {I_r^0,I_r^1, \cdots ,I_r^{L - 1}} \right\}$ and $\left\{ {I_t^0,I_t^1, \cdots ,I_t^{L - 1}} \right\}$, respectively, where $I_r^0 = {I_r}$ is the ``zero$^{th}$" level image, $I_r^n$ is downsampled from $I_r^{n - 1}$ and the resolution of $I_r^n$ is half of $I_r^{n - 1}$. Then, we implement SfMNeRF from the lowest level images $I_r^{L - 1}$ and $I_t^{L - 1}$, and obtain the depth map of the reference image $I_r^{L - 1}$ as $D_r^{L - 1}$ which served as the depth prior for $I_r^{L - 2}$; we then implement SfMNeRF again with $I_r^{L - 2}$ and $I_t^{L - 2}$ as the input, however, a depth loss is added in the training loss to take advantage of the depth prior estimated in the last step. The depth loss is defined as

\begin{equation}\label{eq16}
{L_{depth}} = \sum\limits_{p\left( k \right) \in {\Omega _p}} {\left| {D\left( {p\left( k \right)} \right) - D_r^{L - 1}\left( {p\left( k \right)/2} \right)} \right|} ,
\end{equation}
where ${D\left( {p\left( k \right)} \right)}$ is the estimated depth of pixel ${p\left( k \right)}$ according to Eq. (7) and ${D_r^{L - 1}\left( {p\left( k \right)/2} \right)}$ depicts that dividing the 2D coordinates of pixel ${p\left( k \right)}$ by 2 and sample its depth value from $D_r^{L - 1}$ through bilinear interpolation as shown in Section IV(B).

We recursively implement SfMNeRF from the lowest level till the ``zero$^{th}$" level, and the trained model from the ``zero$^{th}$" level served as the final model.

\begin{table*}[htb]
\renewcommand\arraystretch{1.2}
\footnotesize
\begin{center}
\begin{tabular}{p{3.5cm}|p{0.85cm}p{0.85cm}p{0.85cm}|p{0.85cm}p{0.85cm}p{0.85cm}|p{0.85cm}p{0.85cm}p{0.85cm}}
\shline
\multirow{2}{*}{Method}  & \multicolumn{3}{c|}{\textbf{PSNR}$\uparrow$} & \multicolumn{3}{c|}{\textbf{SSIM}$\uparrow$} & \multicolumn{3}{c|}{\textbf{LPIPS $_vgg$}$\downarrow$}
\\
\cline{2-10}
& 2-view & 5-view & 8-view & 2-view & 5-view & 8-view & 2-view & 5-view & 8-view \\
\hline
NeRF \cite{Mildenhall2020} ECCV20 & 15.3 &	20.5 &	22.4 &	0.385 &	\textbf{0.708} &	0.773 &	0.537 &	0.273 &	0.226\\
DietNeRF \cite{JainA2021} ICCV21 & 16.4 &	20.7 &	\textbf{22.5} &	0.431 &	0.695 &	0.752 &	0.543 &	0.32 &	0.282\\
Mip-NeRF 360 \cite{Barron2022} CVPR22 & 15.5 &	19.5 &	22.2 &	0.42 &	0.677 &	\textbf{0.78} &	0.469 &	\textbf{0.26} &	\textbf{0.182}\\
FreeNeRF \cite{Yang2023} CVPR23 & \textbf{17.9} &	\textbf{20.9} &	\textbf{22.5} &	\textbf{0.536} &	0.696 &	0.766 &	\textbf{0.364} &	\textbf{0.26} &	0.209\\
Ours  & 16.3 &	20.4 &	22.2 &	0.433 &	0.674 &	0.738 &	0.528 &	0.332 &	0.286\\
\shline

\end{tabular}
\end{center}
\caption{Quantitative comparisons for novel-view synthesis on LLFF-NeRF dataset. Best results shown in bold.}
\end{table*}

\begin{table*}[htb]
\renewcommand\arraystretch{1.2}
\footnotesize
\begin{center}
\begin{tabular}{p{3.5cm}|p{0.85cm}p{0.85cm}p{0.85cm}|p{0.85cm}p{0.85cm}p{0.85cm}|p{0.85cm}p{0.85cm}p{0.85cm}}
\shline
\multirow{2}{*}{Method}  & \multicolumn{3}{c|}{\textbf{PSNR}$\uparrow$} & \multicolumn{3}{c|}{\textbf{SSIM}$\uparrow$} & \multicolumn{3}{c|}{\textbf{LPIPS $_vgg$}$\downarrow$}
\\
\cline{2-10}
& 2-view & 5-view & 8-view & 2-view & 5-view & 8-view & 2-view & 5-view & 8-view \\
\hline
NeRF \cite{Mildenhall2020} ECCV20 & 11.0 &	12.6 &	13.9 &	0.425 &	0.528 &	0.565 &	0.711 &	0.693 &	0.673\\
DietNeRF \cite{JainA2021} ICCV21 & 10.8 &	11.5 &	12.4 &	0.427 &	0.483 &	0.472 &	0.744 &	0.707 &	0.709\\
Mip-NeRF 360 \cite{Barron2022} CVPR22 & 14.6 &	15.8 &	18.3 &	0.456 &	0.531 &	0.616 &	\textbf{0.567} &	\textbf{0.521} &	\textbf{0.455}\\
FreeNeRF \cite{Yang2023} CVPR23 & 12.6 &	14.2 &	15.5 &	0.377 &	0.418 &	0.492 &	0.589 &	0.543 &	0.52\\
Ours  & \textbf{14.8} &	\textbf{17.5} &	\textbf{19.6} &	\textbf{0.503} &	\textbf{0.655} &	\textbf{0.656} &	0.638 &	0.602 &	0.555\\
\shline

\end{tabular}
\end{center}
\caption{Quantitative comparisons for novel-view synthesis on ScanNet dataset. Best results shown in bold.}
\end{table*}

\begin{table*}[htb]
\renewcommand\arraystretch{1.2}
\footnotesize
\begin{center}
\begin{tabular}{p{3.4cm}|p{0.8cm}p{0.8cm}p{0.8cm}|p{0.8cm}p{0.8cm}p{0.8cm}|p{0.8cm}p{0.8cm}p{0.8cm}}
\shline
\multirow{2}{*}{Method}  & \multicolumn{3}{c|}{\textbf{PSNR}$\uparrow$} & \multicolumn{3}{c|}{\textbf{SSIM}$\uparrow$} & \multicolumn{3}{c|}{\textbf{LPIPS $_vgg$}$\downarrow$}
\\
\cline{2-10}
& 2-view & 5-view & 8-view & 2-view & 5-view & 8-view & 2-view & 5-view & 8-view\\
\hline
NeRF \cite{Mildenhall2020} ECCV20 & 10.0 &	10.2 &	10.4 &	0.380 &	0.391 &	0.406 &	0.658 &	0.671 &	0.695\\
DietNeRF \cite{JainA2021} ICCV21 & 9.32 &	9.22 &	9.65 &	0.327 &	0.325 &	0.358 &	0.676 &	0.685 &	0.716\\
Mip-NeRF 360 \cite{Barron2022} CVPR22 & 8.79 &	9.17 &	9.17 &	0.208 &	0.227 &	0.242 &	0.662 &	0.652 &	0.643\\
FreeNeRF \cite{Yang2023} CVPR23 & 8.10 &	8.88 &	9.02 &	0.203 &	0.223 &	0.247 &	0.638 &	0.629 &	0.621\\
Ours  & \textbf{12.8} &	\textbf{16.2} &	\textbf{17.7} &	\textbf{0.453} &	\textbf{0.625} &	\textbf{0.642} &	\textbf{0.55} &	\textbf{0.427} &	\textbf{0.428}\\
\shline

\end{tabular}
\end{center}
\caption{Quantitative comparisons for novel-view synthesis on DTU dataset. Best results shown in bold.}
\end{table*}

\begin{table*}[htb]
\renewcommand\arraystretch{1.2}
\footnotesize
\begin{center}
\begin{tabular}{p{7.0cm}|p{0.8cm}p{0.8cm}p{2.5cm}}
\shline
Methods & \textbf{PSNR}$\uparrow$ & \textbf{SSIM}$\uparrow$ & \textbf{Training Run Time}\\
\hline
\emph{Basic}	& 11.2 & 0.428 &  0.27291 \\
\emph{Basic+3D}	& 11.6	& 0.435 & 0.27365 \\
\emph{Basic+3D+ds}	& 11.8	& 0.436  & 0.27704 \\
\emph{Basic+3D+ds+pr}	& 13.2	& 0.451 & 0.27729 \\
\emph{Basic+3D+ds+pr+SSIM}	& 13.7	& 0.456 & 0.27760 \\
\emph{Basic+3D+ds+pr+SSIM+epi}	& 14.4	& 0.472 & 0.43389 \\
\emph{Basic+3D+ds+pr+SSIM+sub-pixel sampling+epi}	& 14.8	& 0.503 & 0.43589 \\

\shline

\end{tabular}
\end{center}
\caption{Ablation study on SfMNeRF. The evaluation is performed on ScanNet dataset (2 input views).}
\end{table*}

\section{Experiment}
\subsection{Experimental Details}
We employed the commonly used deep learning library Pytorch to implement our approach. The performance of our system is evaluated on the LLFF-NeRF \cite{Mildenhall2020, Mildenhall2019}, ScanNet \cite{DaiA2017} and DTU MVS (DTU) \cite{Jensen2014} datasets.

1)	Datasets: The LLFF Dataset is established from eight scenes captured by a cellphone, with 20-62 images each. The resolution of each image in the dataset is $4032 \times 3024$. We downsized each image 1/8 scale to $504 \times 378$ dimensions in pixels due to the limited capacity of NVIDIA RTX 3090, and held out 1/8 of these as the test set for novel view synthesis.

For the ScanNet dataset, we selected eight scenes in this dataset to evaluate our method as the experimental setup in \cite{Mildenhall2019}. In each scene, 40 images were selected to cover a local region, and all images were resized to $648 \times 484$. As introduced in NeRF \cite{Mildenhall2020}, we held out 1/8 of these as the test set for novel view synthesis.

For the DTU dataset, we adhere to the protocol of Yu et al. \cite{YuA2021}. Yu et al. \cite{YuA2021} reported their experimental results on 15 selected scenes, but we only evaluate on 13 scenes since COLMAP \cite{Schonberger2016} fail to estimate the camera poses for scenes scan40 and scan103. Each image in the test set was downsized 1/4 scale to a resolution of $400 \times 300$ .

Following previous work \cite{YuA2021, Niemeyer2022}, the experimental results were evaluated on 2, 5, and 8 input views, respectively. The input views were selected from the neighbourhood of the mid image in the train set.

2)	Training Details: We set ${\lambda _{ren}} = 1.0,{\rm{ }}{\lambda _{3D}} = 1.0,{\rm{ }}{\lambda _{pr}} = 0.001,{\lambda _{epi}} = 0.0001,{\rm{ }}{\lambda _{SSIM}} = 0.008$ and ${\lambda _{ds}} = 0.01$ in all the scenes of each dataset. For fair comparison, the same MLP architecture in NeRF \cite{Mildenhall2020} was employed in all experiments. The patch size was set as $32 \times 32$. The hierarchical sampling strategy in NeRF \cite{Mildenhall2020} was adopted and numbers of sampled points of both coarse sampling and importance sampling were set to 64. We use the Adam \cite{Kingma2014} optimizer to optimize our model with ${\beta _1} = 0.9,{\beta _2} = 0.999$ and the learning rate was set as 1e-3. In all experiments, we used one NVIDIA RTX 3090 to training and testing.

3)	Evaluation Metrics: We employed three kinds of metrics to evaluate the quality of novel-view rendering: Peak Signal-to-Noise Ratio (PSNR), Structural Similarity Index Measure (SSIM) \cite{WangZ2004} and Learned Perceptual Image Patch Similarity (LPIPS) \cite{ZhangR2018}.

\subsection{Comparison with State-of-the-Art}
For all datasets, we compare quantitatively and quantitatively to the original approach (NeRF) \cite{Mildenhall2020} and three recently proposed works in sparse-inputs novel view synthesis: DietNeRF \cite{JainA2021}, Mip-NeRF 360 \cite{Barron2022} and FreeNeRF \cite{Yang2023}.

\noindent \textbf{Results on LLFF-NeRF dataset.}

In our study, we conducted a comparative analysis of SfMNeRF against NeRF \cite{Mildenhall2020} and three recently proposed methodologies. The quantitative assessments for novel-view synthesis are delineated in Table I, and the visual outcomes are depicted in Fig. 4. Examination of Table I reveals that SfMNeRF surpasses NeRF and Mip-NeRF 360 in terms of PSNR and SSIM when utilizing two input views. However, our approach yields inferior performance in other metrics, attributed to the presence of rich textures in the LLFF-NeRF dataset-essential for NeRF and its derivatives, but not fully adhered to by our methodology. Fig. 4 provides visual evidence that, for two input views, our approach adeptly predicts high-quality novel views in comparison to alternative methods.

\noindent \textbf{Results on ScanNet dataset.}

Quantitative assessments for novel-view synthesis on the ScanNet dataset are outlined in Table II. Our approach demonstrates superior results measured by PSNR and SSIM across two, five, and eight input views. Although our approach lags in LPIPS performance, it's important to note that this metric is not directly aligned with the objective of novel view synthesis, as highlighted by other works. Fig. 5 presents qualitative results, revealing that novel views synthesized by other methods often lack recognizability. Even with five or eight input views, these results exhibit floating artifacts. Leveraging geometric constraints within multi-views, our approach consistently performs well across all scenarios.

\noindent \textbf{Results on DTU dataset.}

Evaluations on the DTU dataset, detailed in Table III, demonstrate the significant superiority of our approach over other methodologies, particularly when only five and eight input views are available. Qualitative results in Fig. 6 illustrate that other methods struggle to synthesize plausible novel views. Artifacts present in views generated by NeRF and DietNeRF are attributed to photometric ambiguity in texture-less regions, challenging for NeRF to address. Conversely, our approach accurately synthesizes novel views even with only two input views, showcasing its ability to effectively tackle such ambiguity through a comprehensive solution. The corresponding estimated depth maps are illustrated in Fig.7, SfMNeRF successfully recovered the accurate 3D-scene representation across all views. In contrast, the other methods failed to reconstruct the corrected 3D-scene structure, which in turn deteriorated the view synthesis.

\subsection{Ablation Study}
We conducted extensive ablation studies on the ScanNet dataset to validate the effectiveness of the individual components in SfMNeRF for view synthesis. Table IV shows the quantitative results. We explain items in Table IV as follows:

\emph{Basic}: Only the rendering loss is used.

\emph{3D}:  The positions-of-matched-features loss is included.

\emph{ds}:  The depth smooth loss is included.

\emph{pr}:  The photometric reconstruction loss between patches in different views is included.

\emph{SSIM}: The SSIM loss between patches in different views is included.

\emph{sub-pixel sampling}: The sub-pixel sampling strategy is implemented during training.

\emph{epi}: The epipolar constraint is included.

From Table IV, we observe that each loss employed in SfMNeRF is able to improve novel-view synthesis accuracy. The average training run times for the individual components are also reported, as illustrated in column 3 of Table IV. All values are denoted in milliseconds. The training run time is measured by the average time to train a batch. We have observed that each loss contributes to an increase in training run time, with the epipolar constraint being particularly time-consuming.

\section{Conclusion and Future Work}
In this paper, we analyzed the limitation of NeRF and proposed SfMNeRF, a neural radiance field is able to improve the quality of novel view synthesis by self-supervised depth constraints. The employed depth priors are obtained by some constraints, such as patch-based photometric consistency constraint between multiple views, epipolar constraint and positions-of-matched-features constraint, without additional data. In this way, our approach learns a multi-view consistent geometry with depth constraints. The depth priors can eliminate the geometric ambiguity to some extent and improve the quality of novel-view synthesis accordingly. The employed sub-pixel sampling strategy introduces more samples in training that further implicitly constrain the 3D-scene geometry in NeRF. By employing explicit and implicit depth constraints, our approach improves the novel-view rendering quality of NeRF. In terms of comparison to other depth priors based NeRF, our approach does not acquire to estimate the sparse depth map by a SfM in advance. Our approach motivates the future research to further exploit the structural priors in multi-view inputs for view synthesis and other related tasks.

As with other photometry-based reconstruction methods, SfMNeRF suffers from the scenes with repeated structures which are likely to be caused by photometric ambiguity. In the future, we will investigate how to incorporate the rich priors of indoor datasets, and adapt the proposed approach to achieve a generalized NeRF trained on across large scale datasets.

\section*{Acknowledgment}

This research is supported in part by the the Research Foundation of Education Bureau of Hunan Province (No. 22A0124) and National Key R\&D Program of China (No. 2018AAA0102102).

\ifCLASSOPTIONcaptionsoff
  \newpage
\fi

\end{document}